\documentclass[runningheads]{llncs}
\if{CLASSINFOpdf}
   \usepackage[pdftex]{graphicx}
\else
   \usepackage[dvips]{graphicx}
\fi

\usepackage[linesnumbered,ruled]{algorithm2e}
\usepackage[cmex10]{amsmath}
\usepackage{array}
\usepackage{fixltx2e}
\usepackage{enumerate}
\usepackage{multicol}
\usepackage{multirow}
\usepackage{mathtools}

\DeclarePairedDelimiter{\norm}{\lVert}{\rVert}
\usepackage{mathptmx} 
\usepackage{threeparttable}
\usepackage{cite}
\usepackage{array}
\usepackage{color, colortbl}
\definecolor{Gray}{gray}{0.9}
\definecolor{LightCyan}{rgb}{0.88,1,1}
\usepackage{amssymb}
\usepackage{amsfonts}
\usepackage{amsmath}
\usepackage{verbatim}
\usepackage{urwchancal, pifont}
\usepackage{bbm}
\usepackage{cancel}

\newtheorem{prop}{Proposition}

\usepackage{color}
\usepackage{subcaption, url}
\newcommand*{\nindep}{%
  \mathbin{
    \mathpalette{\@indep}{\not}
  }%
}

\makeatletter
\def\BState{\State\hskip-\ALG@thistlm}
\makeatother

\begin{document}

\title{Communal Domain Learning for Registration in Drifted Image Spaces}
\titlerunning{Communal Domain Learning for Registration in Drifted Image Spaces}
\author{Awais Mansoor\inst{1,2}\orcidID{0000-0001-6761-8030}
\thanks{Corresponding author: awais.mansoor@gmail.com.}
\and Marius George Linguraru\inst{1}
}
\institute{$^1$The Sheikh Zayed Institute for Pediatric Surgical Innovation, Children's National Health System, Washington DC.\\
$^2$Digital Technology and Innovation, Siemens Healthineers, Princeton, NJ, USA.}

\authorrunning{A. Mansoor et al.}
%


\maketitle

\begin{abstract}
Designing a registration framework for images that do not share the same probability distribution is a major challenge in modern image analytics yet trivial task for the human visual system (HVS). Discrepancies in probability distributions, also known as \emph{drifts}, can occur due to various reasons including, but not limited to differences in sequences and modalities (e.g., MRI T1-T2 and MRI-CT registration), or acquisition settings (e.g., multisite, inter-subject, or intra-subject registrations). The popular assumption about the working of HVS is that it exploits a communal feature subspace exists between the registering images or fields-of-view that encompasses key drift-invariant features. Mimicking the approach that is potentially adopted by the HVS, herein, we present a representation learning technique of this invariant communal subspace that is shared by registering domains. The proposed communal domain learning (CDL) framework uses a set of hierarchical nonlinear transforms to learn the communal subspace that minimizes the probability differences and maximizes the amount of shared information between the registering domains. Similarity metric and parameter optimization calculations for registration are subsequently performed in the drift-minimized learned communal subspace. This generic registration framework is applied to register multisequence (MR: T1, T2) and multimodal (MR, CT) images. Results demonstrated generic applicability, consistent performance, and statistically significant improvement for both multi-sequence and multi-modal data using the proposed approach ($p$-value$<0.001$; Wilcoxon rank sum test) over baseline methods.
\keywords{Image registration, Multimodal images, Multisequence images, Domain adaptation.}
\end{abstract}

\section{Introduction}
\vspace{-0.1in}
Image registration is a fundamental operation in image-based analytics involving fusion of information, quantitative comparison, or spatial normalization. A typical registration pipeline consists of the following three major components. A \textbf{transformation model ($\mathbf{T}_\mu$)} that defines the geometric relationship between the registering images. Depending upon the parameter vector $\mu$, the transformation model can be rigid, affine, or nonrigid (deformable). A \textbf{similarity metric or cost function ($\mathcal{C}$)} measuring the degree of alignment between the $d$-dimensional image belonging to the target domain ($\mathbf{x}_t\in\mathbb{R}^d$) and the transformed image belonging to the source domain ($\mathbf{T}_\mu\mathbf{x}_s\in\mathbb{R}^d$). The source and the target domains can be identical or different. Then an \textbf{optimizer} iteratively improves $\mu$ based on $\mathcal{C}$. Specifically, at a $k^\text{th}$ iteration, the current vector $\mu_k$ is updated by taking a step in the search direction $\mathbf{d}_k\mathbf{\mu}_{k+1}=\mathbf{\mu}_k-a_k\mathbf{d}_k$, where $a_k$ is the scalar step size. A typical registration task involves finding the optimal transformation parameters that maximizes the degree of similarity across the registering images: $\hat{\mu}=\operatorname*{arg\,max}_\mu \mathcal{C}\left(\mathbf{T}_\mu\mathbf{x}_s, \mathbf{x}_t\right)$.
\vspace{-0.3in}
\subsubsection{State-of-the-Art Methods.}
The registration of images belonging to domains with substantial discrepancy, also known as \emph{drift}, remains challenging. Drift in medical images generally occurs due to changes in either the field-of-view (e.g., 2D-3D),  modalities or sequences (e.g., CT-MRI, T1-T2 MRI), or acquisition settings (e.g., data acquired with different protocols). Typical approaches to handle drifts use either information theoretic similarity metrics such as mutual information or normalized cross correlation that maximize the transferability of knowledge between domains \cite{sotiras2013deformable}. In addition, significant research effort has been invested in devising methods for the effective representations of registering domains, more recently using the deep learning approaches \cite{de2019deep, cao2018deep, zheng2018pairwise}. These representation transformation approaches known as \emph{domain adaptation} or \emph{transfer learning} methods range from simplistic techniques such as intensity standardization to more sophisticated feature mapping approaches \cite{klein2007evaluation}. Commonly used domain adaptation methods estimate a representation transformation of one registering domain to imitate the second one as accurately as possible. The principal hypothesis behind these approaches is that by reducing the drift that exists between the two images through transformation, the task of the similarity metric can be made easier, thus resulting in a more accurate registration. However, depending upon the extent of drift, the predictability of one domain from another can be very limited even theoretically\footnote{For maximum mutual information and therefore perfect predictability:\\ $I(\mathbf{x}_s;\mathbf{x}_t)=H(\mathbf{x}_s)-\cancelto{0}{H(\mathbf{x}_s|\mathbf{x}_t)}=H(\mathbf{x}_t)-\cancelto{0}{H(\mathbf{x}_t|\mathbf{x}_s)}$, where $I$ and $H$ denotes mutual information and entropy respectively.} which is the major bottleneck in the performance of these methods. 
\vspace{-0.2in}
\subsubsection{Our Contributions.}
The human visual system (HVS) is able to recognize objects despite tremendous variation in their appearance resulting from variation in view, size, lighting, etc. This ability—known as ``invariant'' object recognition—is central to visual perception, yet its computational underpinnings are not well understood. One prominent theory behind the cognitive neuroscience of visual object recognition suggests that a drift invariant communal space could be created by the HVS instead of mapping information between spaces\cite{zoccolan2009rodent}. 

Mimicking the HVS, instead of learning to imitate one domain from another through domain adaptation, in this work, we propose learning the communal feature subspace between domains. As demonstrated in the next section, a communal subspace $\mathbb{W}$ between registering domains that captures images with the same field-of-view can be guaranteed to exist, at least theoretically. Our paper introduces an approach to estimate $\mathbb{W}$ using training instances from the registering domains, what we call communal domain learning (CDL). Specifically, CDL estimates the $\mathbb{W}$ between source and target domains that consecutively: (I) maximizes the amount of shared information between the source and the target domains; and (II) minimizes the probability differences between the two domains. Subsequently, the similarity metric and transform parameters for the registration are performed in the estimated communal subspace $\widehat{\mathbb{W}}$. We demonstrated the efficacy of the generically applicable CDL on the registration of multisequence (MR: T1, T2) and multimodal (MR, CT) brain images. 
\vspace{-0.15in}
\section{Methods}
\begin{figure}
\centering
\includegraphics[width=0.98\textwidth]{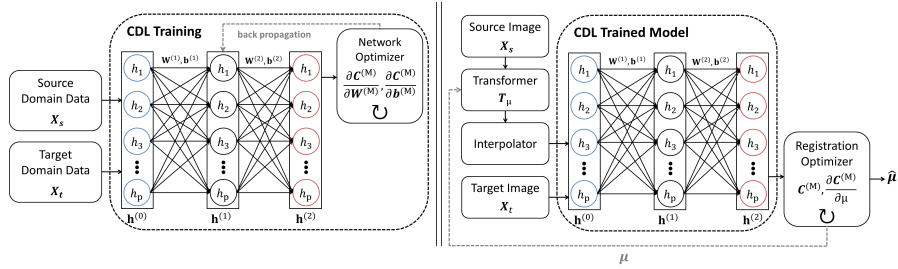}
\caption{\footnotesize{The proposed registration framework with communal domain learning (CDL) network. (\emph{Left}) The CDL network training module. The input to the network are aligned source and target image pairs, output is the learned network parameters $\mathbf{W}^{(m)}$ and $\mathbf{b}^{(m)}$, $1\le m\le 2$. (\emph{Right}) The proposed registration framework using the trained CDL network for similarity calculation of registration hypothesis.}}
\label{fig:flow}
\end{figure}

\vspace{-0.1in}
The flow diagram of the proposed registration framework is presented in Fig. \ref{fig:flow}. The core of the framework is the CDL network: a multi-layered fully connected neural network. CDL learns the communal domain $\mathbb{W}$ between source and target domains through a set of hierarchical nonlinear transformations. During training, CDL takes as input perfectly aligned (through manual inspection, details are explained in the Experiments section) pairs of source and target instances (i.e., $\mathbf{T}_\mu=\mathbbm{1}$). Let $\mathbf{x}=\{(\mathbf{x}_{s_i}, \mathbf{x}_{t_i})|i=1,\dots,N\}$ be the scalar-valued training pair and $N$ is the total number of training tensor pairs. There are $M+1$ layers in the CDL network and $p^{(m)}$ denotes the number of units in the $m^\text{th}$ layer. The output of the network at the $m^\text{th}$ layer is: 
\begin{equation}\mathbf{h}^{(m)}=\phi\left(\mathbf{W}^{(m)}\mathbf{h}^{(m-1)}+\mathbf{b}^{(m)}\right)=\phi\left(\mathbf{z}^{(m)}\right),\label{eq:outputlayer}\end{equation}
\noindent where $\mathbf{W}^{(m)}\in\mathbb{R}^{p^{(m)}\times p^{(m-1)}}$ is the weight matrix and $\mathbf{b}^{(m)}\in\mathbb{R}^{p^{(m)}}$ is the bias vector for the $m^\text{th}$ layer. $\phi$ is the nonlinear activation function and $\mathbf{h}^{(m)}:\mathbb{R}^{p^{(m-1)}}\rightarrow\mathbb{R}^{p^{(m)}}$ is the nonlinear mapping defined over the Hilbert space $\Omega_{\mathbf{h}^{(m)}}$. For the first layer, $\mathbf{h}^{(0)}=\mathbf{x}$ and $p^{(0)}=d$. $d=3$ for volumetric images. 
\vspace{-0.15in}
\subsection{Communal Domain Learning}
\begin{prop}
For sets $\mathbf{x}_s$ and $\mathbf{x}_t$ defined over two separate vector domains $\mathbb{V}_s$ and $\mathbb{V}_t$, respectively, then $\exists\mathbb{W}:\mathbb{W}\subseteq\mathbb{V}_s$ and $\mathbb{W}\subseteq\mathbb{V}_t$ if and only if $\mathbb{V}_s\not\perp\mathbb{V}_t$, where $\mathbb{W}$ is a non-empty subset of vector domains $\mathbb{V}_s$ and $\mathbb{V}_t$.
\label{prop:main}
\end{prop}
\vspace{-0.1in}
In other words, proposition 1 suggests that as long as the two domains ($\mathbb{V}_s$ and $\mathbb{V}_t$) are imaging the same field-of-view, a communal domain ($\mathbb{W}$) possessing drift invariant features can be theoretically guaranteed to exist between them. To learn the communal domain $\mathbb{W}$, it is desired to adjust the CDL network parameters ($\theta=\{\mathbf{W}, \mathbf{b}\}$) that satisfy the constraints I and II, described in the previous section, at the output of the network. Henceforth, the cost function $\mathcal{C}$ at the top layer, $M$, is formulated as the following optimization problem:
\vspace{-0.08in}
\begin{footnotesize}
\begin{eqnarray}
\mathop{{\rm{max}}}\limits_{\mathbf{h}^{(M)}}~\mathcal{C}\left(\mathbf{h}_t^{(M)}, \mathbf{h}_{s}^{(M)}\right)&=I(\mathbf{h}_t^{(M)}; \mathbf{h}_{s}^{(M)})-\alpha D(\mathbf{h}_t^{(M)}, \mathbf{h}_{s}^{(M)})-\beta\sum_{m=1}^M(\norm{\mathbf{W}^{(m)}}^2_F+\norm{\mathbf{b}^{(m)}}^2_2),
\label{eq:optimization00}
\end{eqnarray}
\end{footnotesize}
\vspace{-0.02in}
\noindent where $\alpha>0$ and $\beta>0$ are the regularization parameters, $I(\mathbf{h}_t^{(m)}; \mathbf{h}_{s}^{(m)})$ and $D(\mathbf{h}_t^{(m)}, \mathbf{h}_{s}^{(m)})$ denote the mutual information and the distribution difference distance at the $m^\text{th}$ layer respectively. $\norm{\mathbf{Z}}_F$ is the Frobenius norm of the matrix $\mathbf{Z}$. The typical cost function for registration ($\hat{\mu}=\operatorname*{arg\,max}_\mu \mathcal{C}\left(\mathbf{T}_\mu\mathbf{x}_s, \mathbf{x}_t\right)$) and (\ref{eq:optimization00}) are equivalent owing to the way the registration framework is set up Fig. \ref{fig:flow}. The \emph{back-propagation} algorithm for training the neural network requires computing the derivatives of (\ref{eq:optimization00}) with respect to network parameters $\theta$ while the registration optimizer requires the derivative with respect to transformation model parameters $\mathbf{\mu}$.
\vspace{-0.03in}
\subsubsection{Mutual Information.}
\vspace{-0.2in}
Empirical determination of $I(\mathbf{h}_t^{(M)}; \mathbf{h}_{s}^{(M)})$ in (\ref{eq:optimization00}) requires the estimation of the joint and marginal distributions of the source and target domains. Methods such as the \emph{Parzen}-window \cite{klein2007evaluation} are generally used; however, these window-based methods are computationally complex. Instead, we used a parametric modeling approach in CDL. Specifically, Katyal et al. \cite{katyal2013gaussian} demonstrated that the joint and marginal distributions of voxel intensities from multi-sequence/multi-modal MR/CT images that share the same field-of-view, is Gaussian distributed\footnote{\scriptsize{Optimal parametric models are expected to be dependent on source and target domains and is the topic of our future research.}}. Moreover, through Taylor series expansion that the set of hierarchical transformations presented in eq. (\ref{eq:outputlayer}) when applied to Gaussian data preserve its Gaussianity (proof is provided in the supplementary material). Subsequently, for Gaussian distributed $\mathbf{h}_s^{(m)}$ and $\mathbf{h}_t^{(m)}$, their mutual information is a monotonic function of cross-correlation that is lower-bounded by: $I\left(\mathbf{h}_s^{(m)}; \mathbf{h}_t^{(m)}\right)\geq-\frac{1}{2}\left(1-Corr(\mathbf{h}_t^{(m)},\mathbf{h}_s^{(m)})\right)$. Equality in the equation is achieved, if joint Gaussianity is also assumed \cite{kleeman2011information}. Henceforth, 

\begin{footnotesize}
\begin{equation}
I(\mathbf{h}_t^{(m)},\mathbf{h}_s^{(m)})=-\frac{1}{2}\left(1-\frac{\sum_{i=1}^N{\mathbf{h}_{ti}^{(m)}\mathbf{h}_{si}^{(m)}}}{N\sigma_{t}^{(m)}\sigma_{s}^{(m)}}+\frac{\overline{\mathbf{h}_{t}^{(m)}}.\overline{\mathbf{h}_{s}^{(m)}}}{\sigma_{t}^{(m)}\sigma_{s}^{(m)}}\right),
\label{eq:mutualinformation}
\end{equation}
\end{footnotesize}

\noindent where overbar indicates the expected value. $\sigma_s^{(m)}$ and $\sigma_t^{(m)}$ denote the standard deviation of source and target domain data, respectively. Subsequently, the gradients of mutual information with respect to network parameters $\mathbf{W}^{(m)}$ and $\mathbf{b}^{(m)}$ are computed as follows:
\begin{gather}
\frac{\partial}{\partial \mathbf{W}^{(m)}}I(\mathbf{h}_t^{(m)},\mathbf{h}_s^{(m)})=\left(N\sigma_{\mathbf{h}_t}^{(m)}\sigma_{\mathbf{h}_s}^{(m)}\right)^{-1}\sum_{i=1}^{N}{\left(\mathbf{L}_{ti}^{(m)}\mathbf{h}^{(m-1)}_{ti}+\mathbf{L}_{si}^{(m)}\mathbf{h}^{(m-1)}_{si}\right)},
\label{eq:miW}\\
\frac{\partial}{\partial \mathbf{b}^{(m)}}I(\mathbf{h}_t^{(m)},\mathbf{h}_s^{(m)})=\left(N\sigma_{\mathbf{h}_t}^{(m)}\sigma_{\mathbf{h}_s}^{(m)}\right)^{-1}\sum_{i=1}^{N}{\left(\mathbf{L}_{ti}^{(m)}+\mathbf{L}_{si}^{(m)}\right)},
\label{eq:miB}
\end{gather}
\vspace{-0.1in}
\noindent The updating equations in (\ref{eq:miW}) and (\ref{eq:miB}) for back propagation are calculated as:

\begin{minipage}{0.47\textwidth}
\begin{gather}
\mathbf{L}_{ti}^{(M)}=\phi'(\mathbf{z}_{ti}^{(M)})\odot\phi(\mathbf{z}_{si}^{(M)}),\nonumber\\
\mathbf{L}_{ti}^{(m)}=\left(\mathbf{W}^{(m+1)^T}\mathbf{L}_{ti}^{(m+1)}\right),\nonumber
\end{gather}
\end{minipage}
\begin{minipage}{0.47\textwidth}
\begin{gather}
\mathbf{L}_{si}^{(M)}=\phi'(\mathbf{z}_{si}^{(M)})\odot\phi(\mathbf{z}_{ti}^{(M)}),\nonumber\\
\mathbf{L}_{si}^{(m)}=\left(\mathbf{W}^{(m+1)^T}\mathbf{L}_{si}^{(m+1)}\right),\nonumber
\end{gather}
\end{minipage}

\noindent where $\odot$ denotes the element-wise multiplication.
\vspace{-0.1in}
\subsubsection{Distribution Difference.} To estimate the distribution differences between two domains, we apply the maximum mean discrepancy (MMD) criteria \cite{gretton2006kernel}. MMD is a statistical measure that estimates the dependence of two random variables. Henceforth, the distribution difference distance at the $m^\text{th}$ layer is defined as: \begin{equation}D\left(\mathbf{h}_{t}^{(m)}, \mathbf{h}_{s}^{(m)}\right)=\norm{\frac{1}{N}\sum_{i=1}^N{\left(\mathbf{h}_{ti}^{(m)}-\mathbf{h}_{si}^{(m)}\right)}}^2_\text{2}.\label{eq:mmd}\end{equation} Similar to mutual information, the partial derivatives of $D\left(\mathbf{h}_{t}^{(m)}, \mathbf{h}_{s}^{(m)}\right)$ with respect to network parameters are:
\begin{eqnarray}
\frac{\partial}{\partial\mathbf{W}^{(m)}}D\left(\mathbf{h}_{t}^{(m)}, \mathbf{h}_{s}^{(m)}\right)&=&\frac{2}{N}\sum_{i=1}^N(\mathbf{L}_{ti}^{(m)}\mathbf{h}_{ti}^{(m-1)^T}+\mathbf{L}_{si}^{(m)}\mathbf{h}_{si}^{(m-1)^T}),\label{eq:sdW}\\
\frac{\partial}{\partial\mathbf{b}^{(m)}}D\left(\mathbf{h}_{t}^{(m)}, \mathbf{h}_{s}^{(m)}\right)&=&\frac{2}{N}\sum_{i=1}^N(\mathbf{L}_{ti}^{(m)}+\mathbf{L}_{si}^{(m)}),
\label{eq:sdB}
\end{eqnarray}
Please note that although the same symbols are used to denote losses as (\ref{eq:miW}) and (\ref{eq:miB}), they are defined differently below. Subsequently, the updating equations in (\ref{eq:sdW}) and (\ref{eq:sdB}) for the back propagation framework are:

\begin{minipage}{0.47\textwidth}
\begin{eqnarray}
\mathbf{L}_{ti}^{(M)}=\frac{1}{N}\sum_{j=1}^N{\left(\mathbf{h}_{tj}^{(M)}-\mathbf{h}_{sj}^{(M)}\right)}\odot\phi'\left(\mathbf{z}_{ti}^{(M)}\right),\nonumber\\
\mathbf{L}_{si}^{(M)}=\frac{1}{N}\sum_{j=1}^N{\left(\mathbf{h}_{sj}^{(M)}-\mathbf{h}_{tj}^{(M)}\right)}\odot\phi'\left(\mathbf{z}_{si}^{(M)}\right),\nonumber
\end{eqnarray}
\end{minipage}
\begin{minipage}{0.47\textwidth}
\begin{eqnarray}
\mathbf{L}_{ti}^{(m)}=\left(\mathbf{W}^{(m+1)^T}\mathbf{L}_{ti}^{(m+1)}\right),\nonumber\\
\mathbf{L}_{si}^{(m)}=\left(\mathbf{W}^{(m+1)^T}\mathbf{L}_{si}^{(m+1)}\right).\nonumber
\end{eqnarray}
\end{minipage}

\hspace{0.9\textwidth}

\noindent Algorithm \ref{alg:dITML} summarizes the training of the CDL network (Fig. \ref{fig:flow}(\emph{Left})). The final form of the gradients of the cost function in (\ref{eq:optimization00}) with respect to $\mathbf{W}^{(m)}$ and $\mathbf{b}^{(m)}$ are also provided in (\ref{eq:weightDerivative}) and (\ref{eq:biasDerivative}), respectively, in Algorithm \ref{alg:dITML}.
\begin{algorithm}[htb]
\begin{scriptsize}
    \SetKwInOut{Input}{Input}
    \SetKwInOut{Output}{Output}    
    \Input{Pair-wise training data ($\mathbf{x}$). Free parameters $\alpha$, $\beta$; learning rate $\lambda$; convergence error $\varepsilon$, and maximum number of iterations $K$.}
    \Output{Weights $\{\mathbf{W}^{(m)}\}_{m=1}^M$ and biases $\{\mathbf{b}^{(m)}\}_{m=1}^M$ after convergence.}    
		
		\For{$k=1,\dots,T$}
      {			
				Perform forward-propagation\;
        Compute mutual information (\ref{eq:mutualinformation}) and maximum mean discrepancy (\ref{eq:mmd})\;								
				\For{$m=M-1,\dots,1$}
				{
					Calculate
					\vspace{-0.1in}
					\begin{equation}\begin{split}\frac{\partial}{\partial \mathbf{W}^{(m)}}\mathcal{C}&=-\frac{1}{2}\left(N\sigma_{\mathbf{h}_t}^{(m)}\sigma_{\mathbf{h}_s}^{(m)}\right)^{-1}\sum_{i=1}^N{\left(\mathbf{L}_{ti}^{(m)}\mathbf{h}^{(m-1)}_{ti}+\mathbf{L}_{si}^{(m)}\mathbf{h}^{(m-1)}_{si}\right)}-\frac{2\alpha}{N}\sum_{i=1}^N(\mathbf{L}_{ti}^{(m)}\mathbf{h}_{ti}^{(m-1)^T}\\&+\mathbf{L}_{si}^{(m)}\mathbf{h}_{si}^{(m-1)^T})-2\beta\mathbf{W}^{(m)}\end{split}\label{eq:weightDerivative}\end{equation}\vspace{-0.3in}					 
\begin{equation}\frac{\partial}{\partial \mathbf{b}^{(m)}}\mathcal{C}=-\frac{1}{2}\left(N\sigma_{\mathbf{h}_t}^{(m)}\sigma_{\mathbf{h}_s}^{(m)}\right)^{-1}\sum_{i=1}^N{\left(\mathbf{L}_{ti}^{(m)}+\mathbf{L}_{si}^{(m)}\right)}-\frac{2\alpha}{N}\sum_{i=1}^N(\mathbf{L}_{ti}^{(m)}+\mathbf{L}_{si}^{(m)})-2\beta\mathbf{b}^{(m)}\label{eq:biasDerivative}\end{equation} 						
					using back-propagation.
				
				}			
			\For{$m=1,\dots,M$}
      {
        $\mathbf{W}^{(m)}\leftarrow\mathbf{W}^{(m)}-\lambda\frac{\partial}{\partial \mathbf{W}^{(m)}}\mathcal{C}$;\tcp{Iteratively update weights.}
				$\mathbf{b}^{(m)}\leftarrow\mathbf{b}^{(m)}-\lambda\frac{\partial}{\partial \mathbf{b}^{(m)}}\mathcal{C}$;\tcp{Iteratively update biases.}
      }
			$\lambda\leftarrow 0.95\times\lambda$;\tcp{Reduce the learning rate.}
			Calculate $\mathcal{C}_k$\;
			\If{$\left(|\mathcal{C}_k-\mathcal{C}_{k-1}|<\varepsilon\right)\lor\left(k\ge K\right)$}
			{\Return $\{\mathbf{W}^{(m)}\}_{m=1}^M$ and $\{\mathbf{b}^{(m)}\}_{m=1}^M$;\tcp{Network optimization.}
			}			
      }				
\end{scriptsize}
    \caption{Training of Communal Domain Learning (CDL) Network.}
		\label{alg:dITML}
\end{algorithm}
\vspace{-0.1in}
\subsection{Registration Parameter Estimation}
\vspace{-0.05in}
The registration framework searches for the optimal transformation model parameters in the communal subset domain by performing a constrained hypothesis search within the valid parameter space \cite{klein2007evaluation}. Although several constrained hypothesis strategies have been adopted in the literature \cite{klein2007evaluation}, the commonality is the expression for search direction $\mathbf{d}_k\propto\partial\mathcal{C}/\partial \mathbf{\mu}_k$. Therefore, differentiating $\mathcal{C}$ at the output of the network with respect to the parameter vector at the $k^\text{th}$ iteration yields the search direction:
\begin{equation}
\mathbf{d}_k=\mathbf{h}_t^{(M)}\phi'\left(\mathbf{z}_s^{(M)}\right)\mathbf{W}^{(M)}\frac{\partial}{\partial \mathbf{\mu}_k}\mathbf{T}_\mu-2\alpha\left(\mathbf{h}_{t}^{(M)}-\mathbf{h}_{s,\mu}^{(M)}\right)\phi'\left(\mathbf{z}_s^{(M)}\right)\mathbf{W}^{(M)}\frac{\partial}{\partial \mathbf{\mu}_k}\mathbf{T}_\mu,
\end{equation}
where $\mathbf{z}_s^{(m)}=\mathbf{W}^{(m)}\mathbf{h}_{s,\mu}^{(m-1)}+\mathbf{b}^{(m)}$. The optimization strategy ($\mathbf{\mu}_{k+1}=\mathbf{\mu}_k-a_k\mathbf{d}_k$) is subsequently used to iteratively estimate the parameter vector $\mathbf{\mu}$ until convergence, i.e., 
\begin{equation}
\mathop{{\rm{max}}}\limits_{\mathbf{\mu}}~\mathcal{C}=I(\mathbf{h}_t^{(M)}; \mathbf{h}_{s,\mu}^{(M)})-\alpha D(\mathbf{h}_t^{(M)}, \mathbf{h}_{s,\mu}^{(M)})-\beta\sum_{m=1}^M(\norm{\mathbf{W}^{(m)}}^2_F+\norm{\mathbf{b}^{(m)}}^2_2),
\label{eq:optimization01}
\end{equation}
\noindent Since the transformed source image $\left(\mathbf{T_\mu}\mathbf{x}_s\right)$ is also evaluated at non-voxel positions, \emph{B}-spline interpolation to estimate values at voxel locations.
\vspace{-0.15in}
\section{Experiments}
\vspace{-0.1in}
\subsubsection{Data.} \textbf{[MRI]} We used publicly available T1-T2 volumes from IXI dataset (\url{http://brain-development.org/ixi-dataset}) for training. A total of 30 T1-T2 pairs from the dataset were used: 20 pairs for training and the rest for validation. The MIPAV application (\url{https://mipav.cit.nih.gov}) was used for adjustments in alignment followed by expert inspection for training. For testing, we used 53 T1-T2 pairs acquired at our institution. Spatial resolution in the test data ranges from $0.4\text{mm}\times 0.4\text{mm}\times 0.6\text{mm}$ for T1 and $0.43\text{mm}\times 0.43\text{mm}\times (1.2\text{mm}\text{--}4.0\text{mm})$ for T2.\\ \textbf{[CT]} A total of 20 MR (T1)-CT pairs were acquired for training and 12 for validation. Testing was performed on separate 10 pairs. Spatial resolution was $0.4\text{mm}\times 0.4\text{mm}\times 0.6\text{mm}$ for T1 scans and $0.48\text{mm}\times 0.48\text{mm}\times (0.62\text{mm}\text{--}4.0\text{mm})$ for CT.\\
\textbf{Baseline Method.} The baseline method used for comparison purposes is mutual information (MI), the standard metric for multimodal registration. MI-based registration tend to perform better when image domains are restricted to the object of interest \cite{navab2016deep}. Therefore, we used a fixed intensity threshold of $0.01$ for masking the background; we denote this variant by MI+M. Furthermore, to restrict to the whole brain region, brain extraction tool (BET) \cite{jenkinson2005bet2} was used to obtain the whole brain mask; this variant is denoted by MI+B.  Unfortunately, we could not compare our approach with other learning-based metrics \cite{de2019deep, cao2018deep, zheng2018pairwise, navab2016deep} as their implementation was not available.\\
\textbf{Miscellaneous Implementation Details.} We used Theano with Keras wrapper, Nvidia Titan X GPU, CUDA 7.5, and CuDNN 4.0 for network training. Sigmoid activation, $\lambda=0.2$, $\alpha=0.1$, and $\beta=10$ were used in CDL network. The registration pipeline for every framework consists of regular step gradient decent (max 500 iterations, step size $a_k=0.2/k$). 75 histogram bins were adopted for MI, MI+M, and MI+B variants.
\vspace{-0.1in}
\subsection{Performance}
\vspace{-0.1in}
\subsubsection{Registration Accuracy.} Statistically significant improvement in the registration performance was observed for both multi-sequence and multi-modal data using the proposed approach ($p$-value$<0.001$; Wilcoxon rank sum test) (Table \ref{table:results}). Masking had negligible effect on the performance of the proposed method.
\vspace{-0.2in}
\subsubsection{Plausibility of Cost Function and Search Direction.} To investigate the gain of registering in the communal domain, we monitor the behavior of $\mathcal{C}$ and its derivative in determining the optimal search direction. We randomly perturbed a rotation parameter of the transformation for one image per pair using the 10 aligned validation pairs. Fig. \ref{fig:results}(a) presents the scatter plots of initial and final Dice scores for each registration pair. The scatter plot demonstrates superior performance of the proposed method for all registering pairs. Fig. \ref{fig:results}(b) shows the average gain in Dice score in each iteration. Along with Fig. \ref{fig:results}(a), the plot also demonstrates faster convergence.
\vspace{-0.2in}
\begin{table}
\centering
\begin{scriptsize}
    \begin{tabular}{| p{3cm} | c | c | c | c |c | c |}
    \hline
		&&&\multicolumn{4}{c|}{\textbf{Mask}}\\
		\cline{4-7}
		&&&\multicolumn{2}{c|}{\textbf{Background Subtracted}}&\multicolumn{2}{c|}{\textbf{Whole Brain}}\\
		\hline
		&\multicolumn{ 2}{c|}{\multirow{ 2}{*}{\textbf{Method}}}&\multirow{ 2}{*}{\textbf{Dice}}&\textbf{HD}
		&\multirow{ 2}{*}{\textbf{Dice}}&\textbf{HD}\\
		&&&\textbf{Score}&\textbf{(mm)}&\textbf{Score}&\textbf{(mm)}\\
    \hline
		\hline
    \multirow{2}{*}{\textbf{Multisequence}}&\multirow{3}{*}{MI} & None & $0.80\pm0.05$ & $122.41\pm37.16$&$0.76\pm0.54$&$87.41\pm25.33$\\ \cline{3-7}
     && MI+M & $0.86\pm0.17$ & $107.78\pm29.21$&$0.80\pm0.27$&$76.15\pm18.91$\\ \cline{3-7}
     \multirow{2}{*}{\textbf{Registration (T1-T2)}}&& MI+B & N/A & N/A&$0.82\pm0.21$&$58.11\pm16.77$\\ \cline{2-7}   		
     &\textbf{Proposed}&\textbf{None}&$\mathbf{0.93\pm0.18}$&$\mathbf{91.77\pm18.12}$&$\mathbf{0.86\pm0.19}$&$\mathbf{55.95\pm19.09}$ \\
    \hline		
		\hline
    \multirow{2}{*}{\textbf{Multimodal}}&\multirow{2}{*}{MI} & None & $0.83\pm0.11$ & $108.33\pm11.21$&N/A&N/A\\ \cline{3-7}
     && MI+M & $0.89\pm0.21$ & $98.79\pm30.11$&N/A&N/A\\ \cline{2-7}        		
     \textbf{Registration (T1-CT)}&\textbf{Proposed}&\textbf{None}&$\mathbf{0.96\pm0.26}$&$\mathbf{82.21\pm31.18}$&N/A&N/A\\
    \hline				
    \end{tabular}
		\end{scriptsize}
		\vspace{0.1in}
		\caption{\footnotesize{Quantitative comparison (Dice score and Hausdorff distance-HD) of multisequence/multimodal affine registration. Please note that the boundaries of whole brain are not clearly imaged in CT hence results for whole brain are not listed for multi-sequence registration.}}
		\label{table:results}
\end{table}
\begin{figure}
\centering
\begin{subfigure}{0.44\textwidth}
\includegraphics[width=\textwidth]{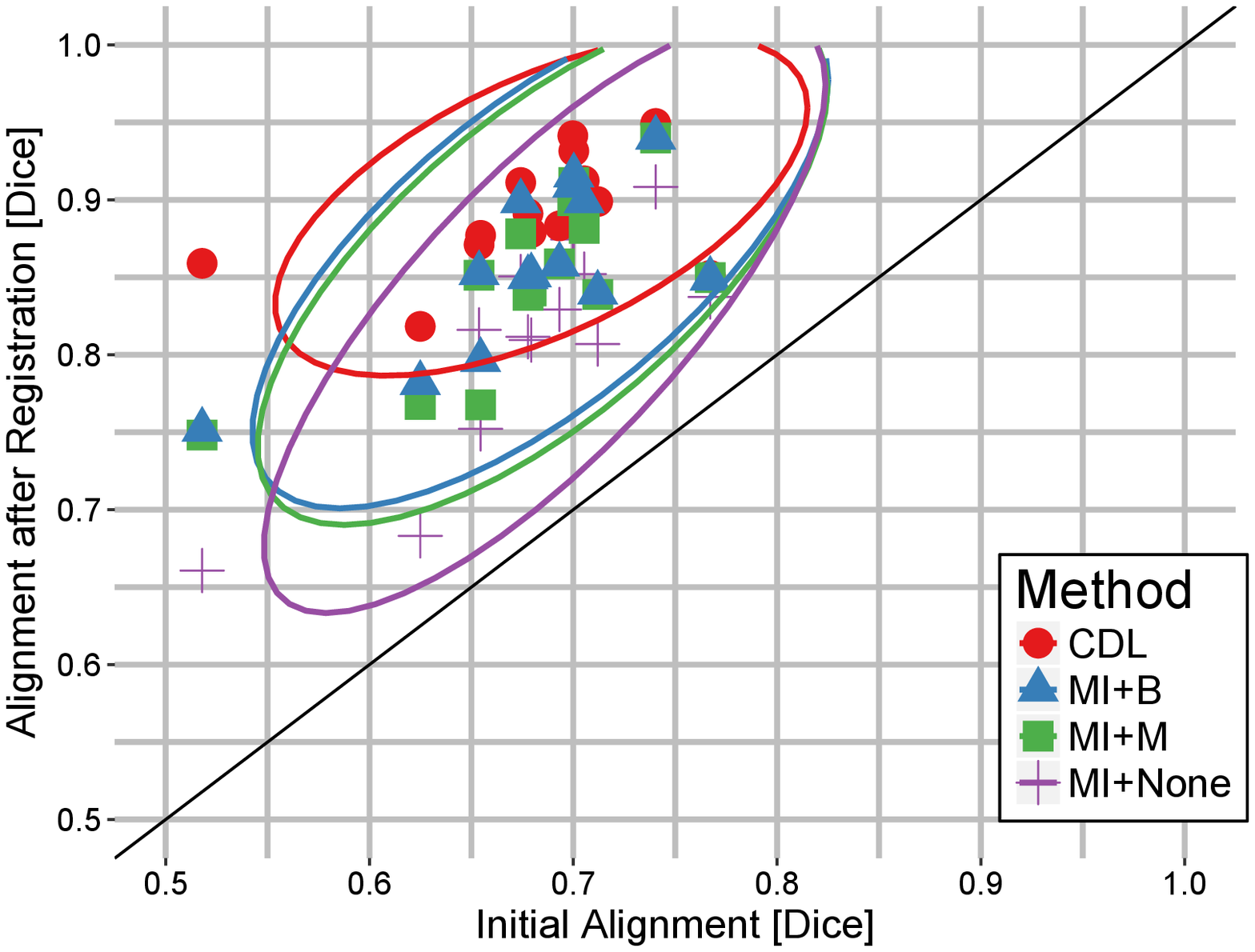}
\caption{}
\end{subfigure}
\begin{subfigure}{0.44\textwidth}
\includegraphics[width=\textwidth]{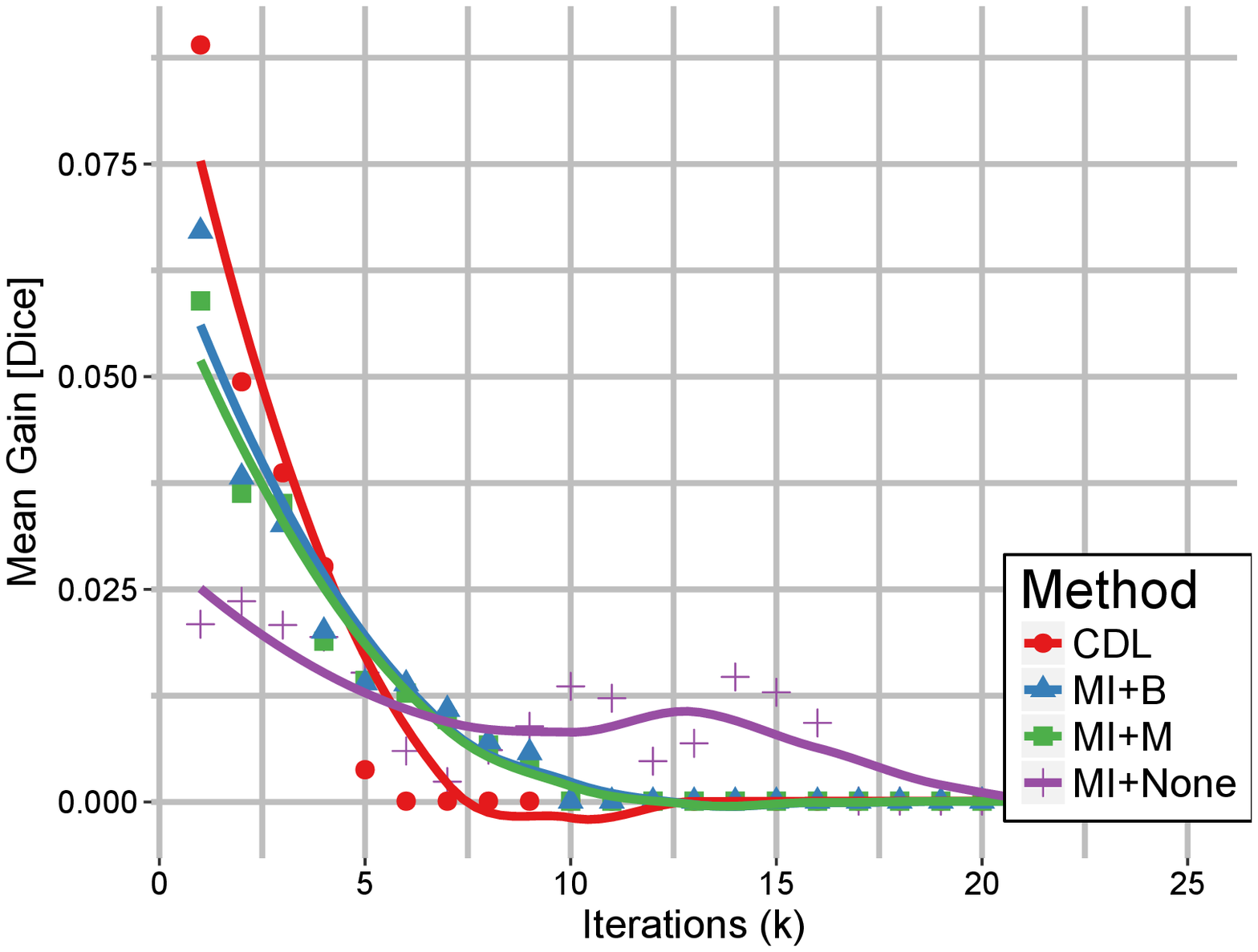}
\caption{}
\end{subfigure}
\caption{\footnotesize{Quantitative performance evaluation of our framework, (a) Dice score gain through registration for the validation data. Each data point represents a registration run; higher is the centroid of the cluster, the greater is the overall improvement. Diagonal line denotes the identity transform ($\mathbf{T}_\mu=\mathbbm{1}$). (b) Mean gain in the Dice score per iteration.}}
\label{fig:results}
\end{figure}
\vspace{-0.65in}
\section{Conclusion}
\vspace{-0.1in}
We introduced a framework for the registration of images that do not share the same probability distribution. Unlike the conventional approaches that aim at reducing the appearance gap between two domains, the proposed approach learns the communal subset domain. Our approach aims to emulates the human visual system for object recognition in the presence of substantial appearance variation. Unlike previous approaches that focus on addressing a specific aspect of registration through learning (e.g., similarity metric, optimization), our framework presents a complete registration pipeline based on learning. Experimental results have demonstrated higher registration accuracy and potential of generic applicability.
\vspace{-0.2in}
\bibliographystyle{splncs}
\vspace{-0.2in}
\bibliography{similaritybib}

\begin{thebibliography}{10}

\bibitem{sotiras2013deformable}
Sotiras, A., Davatzikos, C., Paragios, N.:
\newblock Deformable medical image registration: A survey.
\newblock IEEE transactions on medical imaging \textbf{32}(7) (2013)  1153

\bibitem{de2019deep}
de~Vos, B.D., Berendsen, F.F., Viergever, M.A., Sokooti, H., Staring, M.,
  I{\v{s}}gum, I.:
\newblock A deep learning framework for unsupervised affine and deformable
  image registration.
\newblock Medical image analysis \textbf{52} (2019)  128--143

\bibitem{cao2018deep}
Cao, X., Yang, J., Wang, L., Xue, Z., Wang, Q., Shen, D.:
\newblock Deep learning based inter-modality image registration supervised by
  intra-modality similarity.
\newblock In: International Workshop on Machine Learning in Medical Imaging,
  Springer (2018)  55--63

\bibitem{zheng2018pairwise}
Zheng, J., Miao, S., Wang, Z.J., Liao, R.:
\newblock Pairwise domain adaptation module for cnn-based 2-d/3-d registration.
\newblock Journal of Medical Imaging \textbf{5}(2) (2018)  021204

\bibitem{klein2007evaluation}
Klein, S., Staring, M., Pluim, J.P.:
\newblock Evaluation of optimization methods for nonrigid medical image
  registration using mutual information and {B}-splines.
\newblock IEEE Transactions on Image Processing \textbf{16}(12) (2007)
  2879--2890

\bibitem{zoccolan2009rodent}
Zoccolan, D., Oertelt, N., DiCarlo, J.J., Cox, D.D.:
\newblock A rodent model for the study of invariant visual object recognition.
\newblock Proceedings of the National Academy of Sciences \textbf{106}(21)
  (2009)  8748--8753

\bibitem{katyal2013gaussian}
Katyal, R., Paneri, S., Kuse, M.:
\newblock Gaussian intensity model with neighborhood cues for fluid-tissue
  categorization of multisequence {MR} brain images.
\newblock Proceedings of the MICCAI Grand Challenge on MR Brain Image
  Segmentation (2013)

\bibitem{kleeman2011information}
Kleeman, R.:
\newblock Information theory and dynamical system predictability.
\newblock Entropy \textbf{13}(3) (2011)  612--649

\bibitem{gretton2006kernel}
Gretton, A., Borgwardt, K.M., Rasch, M., Sch{\"o}lkopf, B., Smola, A.J.:
\newblock A kernel method for the two-sample-problem.
\newblock In: Advances in Neural Information Processing Systems. (2006)
  513--520

\bibitem{navab2016deep}
Simonovsky, M., Guti{\'e}rrez-Becker, B., Mateus, D., Navab, N., Komodakis, N.:
\newblock A deep metric for multimodal registration.
\newblock In: International Conference on Medical Image Computing and
  Computer-Assisted Intervention, Springer (2016)  10--18

\bibitem{jenkinson2005bet2}
Jenkinson, M., Pechaud, M., Smith, S.:
\newblock {BET2}: {MR}-based estimation of brain, skull and scalp surfaces.
\newblock In: Eleventh Annual Meeting of the Organization for Human Brain
  Mapping. Volume~17., Toronto (2005)  167

\end{thebibliography}

\clearpage
\appendix
\section{Supplementary Material: Communal Domain Learning for Registration in Drifted Image Spaces}
\begin{figure}
\centering
\includegraphics[width=0.8\textwidth]{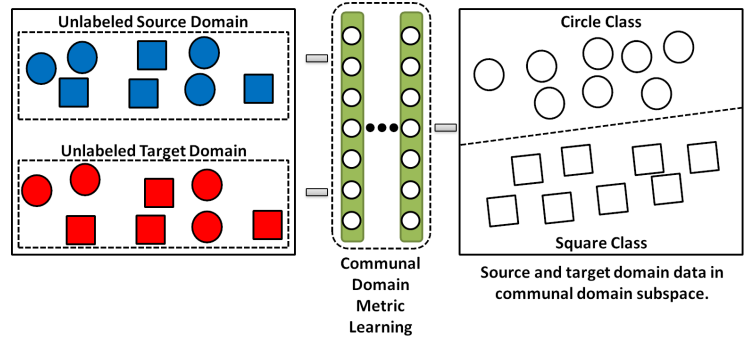}
\caption{\footnotesize{The conceptual diagram of the proposed method. The idea is to obtain the subspace of features from drifted image spaces that are most relevant for the image registration application by getting rid of the features responsible producing drift. For instance, color (blue and red) in this mock example.}}
\label{fig:results}
\end{figure}

\section{Probability density of Gaussian random variables transformed under commonly used activation functions}
Let $\mathbf{Y}=g\left(\mathbf{X}\right)$ be a single-valued continuous transformation of a Gaussian-distributed random variable $\mathbf{X}$. Also let $h\left(\mathbf{Y}\right)$. Then,
\begin{eqnarray}
Pr\left(x_1\le\mathbf{X}\le x_2\right)&=&\int_{x_1}^{x_2}{f\left(x\right)}dx=\int_{y_1}^{y_2}{f\left(h\left(y\right)\right)|h'\left(y\right)|dy},\nonumber\\
&=&Pr\left(y_1\le\mathbf{Y}\le y_2\right)
\label{eq:density01}
\end{eqnarray}
where $f\left(h\left(y\right)\right)|h'\left(y\right)|$ gives the probability density of the transformed random variable $\mathbf{Y}$.
\subsection{The $tanh\left(x\right)$ Activation}
$tanh\left(x\right)$ is a one-to-one function with a continuous first derivative. Then, from eq. (\ref{eq:density01}), the density of $y=tanh\left(x\right)$ is given as: $f\left(y\right)=f\left(h\left(y\right)\right)|h'\left(y\right)|$, where:
\begin{eqnarray}
h\left(y\right)=tanh^{-1}\left(y\right)=\frac{1}{2}\text{ln}\left(\frac{y+1}{y-1}\right)\nonumber
\end{eqnarray}
also $h'\left(y\right)=\frac{1}{1-y^2}$. Using the above results, a random variable $\mathbf{X}\sim N\left(\mu, \sigma^2\right)$ and transformed using $tanh$ activation:
\begin{eqnarray}
f\left(y\right)=\frac{1}{1-y^2}\frac{1}{\sqrt{2\pi\sigma^2}}\exp{\left(-\frac{1}{2\sigma^2}\left[\frac{1}{2}\text{ln}\left(\frac{y+1}{y-1}\right)-\mu\right]^2\right)},
\end{eqnarray}
Furthermore, the Taylor series expansion of is:
\begin{equation}\frac{1}{2}\text{ln}\left(\frac{y+1}{y-1}\right)=y+\frac{y^3}{3}+\frac{y^5}{5}+\dots\end{equation}
From the expansion it is clear that $y=tanh\left(x\right)$ is Gaussian distributed. Also for $y\ll 1$, $\mathbf{Y}\sim N\left(\mu, \sigma^2\right)$.

\subsection{The $sigmoid\left(x\right)$ Activation}
The sigmoid function, $y=sigmoid\left(x\right)=\frac{1}{1+\exp{\left(-x\right)}}$ is also a one-to-one function with a continuous first derivative. Again, from eq. (\ref{eq:density01}), the density of sigmoid function is given as: $f\left(y\right)=f\left(h\left(y\right)\right)|h'\left(y\right)|$, where:
\begin{eqnarray}
h\left(y\right)=\frac{1}{2}\text{ln}\left(\frac{y}{1-y}\right),\nonumber
\end{eqnarray}
and $h'\left(y\right)=\frac{1}{y\left(1-y\right)}$. The above results lead to the fact that for a random variable $\mathbf{X}\sim N\left(\mu, \sigma^2\right)$ that is transformed using $sigmoid\left(x\right)$ activation:
\begin{eqnarray}
f\left(y\right)=\frac{1}{y\left(1-y\right)}\frac{1}{\sqrt{2\pi\sigma^2}}\exp{\left(-\frac{1}{2\sigma^2}\left[\frac{1}{2}\text{ln}\left(\frac{y}{1-y}\right)-\mu\right]^2\right)},
\end{eqnarray}
The Taylor series approximation of $h\left(y\right)$ can be given as:
\begin{equation}\frac{1}{2}\text{ln}\left(\frac{y}{1-y}\right)=y+\frac{y^3}{3}+\frac{y^5}{5}+\dots,\end{equation}
The expansion demonstrates that $y=sigmoid\left(x\right)$ is Gaussian distributed.

\end{document}